\pdfoutput=1

\documentclass[11pt]{article}

\usepackage[final]{acl}

\usepackage{lipsum}
\usepackage{tikz}
\usetikzlibrary{trees,shapes}
\usetikzlibrary{shapes.misc}
\usepackage{xcolor}
\usepackage{epigraph}

\definecolor{hidden-draw}{RGB}{20,68,106}
\definecolor{hidden-pink}{RGB}{255,245,247}
\definecolor{paired-light-yellow}{HTML}{FFFF88}
\definecolor{paired-light-blue}{HTML}{CCE5FF}
\definecolor{paired-light-orange}{HTML}{FFCC99}
\definecolor{paired-dark-yellow}{HTML}{FFF2CC}
\definecolor{paired-light-pink}{HTML}{FFCCCC}
\definecolor{paired-cyan}{HTML}{D5E8D4}
\definecolor{paired-gray}{HTML}{eeeeee}
\definecolor{paired-green}{HTML}{cdeb8b}
\definecolor{paired-blue}{HTML}{dae8fc}
\definecolor{paired-dark-cyan}{HTML}{a2e6eb}
\definecolor{paired-dark-pink}{HTML}{e7b2d2}
\definecolor{paired-purple}{HTML}{9999ff}
\definecolor{paired-pink}{HTML}{cc99ff}
\definecolor{paired-orange}{HTML}{ffcc99}
\definecolor{pic-orange}{HTML}{ffba55}

\definecolor{a1}{RGB}{241,233,191}
\definecolor{a2}{RGB}{255,241,218}

\definecolor{a3}{RGB}{255,239,213}
\definecolor{a4}{RGB}{250,235,215}
\definecolor{a5}{RGB}{255,239,219}
\definecolor{a6}{RGB}{255,246,225}
\definecolor{a7}{RGB}{246,227,201}
\definecolor{a8}{RGB}{254,235,226}
\definecolor{a9}{RGB}{247,220,111}
\definecolor{a10}{RGB}{199,211,189}
\definecolor{a11}{RGB}{209,196,233}
\definecolor{a12}{RGB}{214,234,248}
\definecolor{a13}{RGB}{232,245,233}
\definecolor{a14}{RGB}{237,248,177}
\definecolor{a15}{RGB}{255,228,225}
\definecolor{a16}{RGB}{255,228,181}
\definecolor{a17}{RGB}{255,222,173}
\definecolor{a18}{RGB}{255,218,185}
\definecolor{a19}{RGB}{255,203,164}
\definecolor{a20}{RGB}{247,202,201}

\definecolor{a21}{RGB}{241,254,255}
\definecolor{a22}{RGB}{230,252,252}
\definecolor{a23}{RGB}{179,236,255}
\definecolor{a24}{RGB}{174,226,249}
\definecolor{a25}{RGB}{208,234,246}
\definecolor{a26}{RGB}{189,226,219}
\definecolor{a27}{RGB}{177,204,201}

\definecolor{a28}{RGB}{216,195,216}
\definecolor{a29}{RGB}{195,155,211}
\definecolor{a30}{RGB}{208,152,223}
\definecolor{a31}{RGB}{255,183,209}
\definecolor{a32}{RGB}{255,167,209}
\definecolor{a33}{RGB}{254,235,167}
\definecolor{a34}{RGB}{255,222,137}
\definecolor{a35}{RGB}{254,180,154}
\definecolor{a36}{RGB}{247,148,161}
\definecolor{a37}{RGB}{239,154,154}
\definecolor{a38}{RGB}{255,130,171}
\definecolor{a39}{RGB}{255,105,180}
\definecolor{a40}{RGB}{251,142,172}

\usepackage[edges]{forest}
\usepackage{lipsum}
\usepackage{tikz}
\usetikzlibrary{trees,shapes}

\usepackage{times}
\usepackage{latexsym}
\usepackage{amsmath}
\usepackage{amssymb}
\usepackage{multirow}
\usepackage{bbding}
\usepackage{makecell}
\usepackage{booktabs}
\usepackage{graphicx}
\usepackage{pifont}

\usepackage[T1]{fontenc}

\usepackage[utf8]{inputenc}

\usepackage{microtype}

\usepackage{inconsolata}

\usepackage{graphicx}

\title{Length Extrapolation of Transformers:\\ A Survey from the Perspective of Positional Encoding}

\author{Liang Zhao\textsuperscript{1}, Xiachong Feng\textsuperscript{2}, Xiaocheng Feng\textsuperscript{1,3}\thanks{Corresponding Author}, Weihong Zhong\textsuperscript{1}, \\ {\bf Dongliang Xu\textsuperscript{4}}, {\bf Qing Yang\textsuperscript{4}}, {\bf Hongtao Liu\textsuperscript{4}}, {\bf Bing Qin\textsuperscript{1,3}}, {\bf Ting Liu\textsuperscript{1}}  \\
    \textsuperscript{1}Harbin Institute of Technology \, \textsuperscript{2}The University of Hong Kong \\ \textsuperscript{3}Peng Cheng Laboratory \, 
    \textsuperscript{4}Du Xiaoman Financial (Beijing) \\
    \texttt{\{lzhao, xcfeng, whzhong, qinb, tliu\}@ir.hit.edu.cn}\,
    \texttt{fengxc@hku.hk}\\
    \texttt{\{xudongliang, yangqing, liuhongtao01\}@duxiaoman.com}}

\begin{document}
\maketitle
\begin{abstract}
Built upon the Transformer, large language models (LLMs) have captured worldwide attention due to their remarkable abilities. Nevertheless, all Transformer-based models including LLMs suffer from a preset length limit and can hardly generalize from short training sequences to longer inference ones, namely, they cannot perform \textbf{length extrapolation} to handle long sequences, which severely hinders their application in scenarios demanding long input sequences such as legal or scientific documents. Thus, numerous methods have emerged to enhance the length extrapolation of Transformers. Despite the great research efforts, a systematic survey is still lacking. To fill this gap, we delve into these advances in a unified notation from the perspective of positional encoding (PE),  as it has been considered the primary factor on length extrapolation. Specifically, we begin with extrapolatable PEs that have dominated this research field. Then, we dive into extrapolation methods based on them, covering position interpolation and randomized position methods. Finally, several challenges and future directions in this area are highlighted. Through this survey, we aim to enable the reader to gain a deep understanding of existing methods and provide stimuli for future research.
\end{abstract}

\section{Introduction}
\label{sec:intro}
It has been suggested that with limited learning resources, humans can potentially comprehend utterances of infinite length by understanding their components and structures ~\citep{chomsky1957syntactic, montague1970universal}. In natural language processing (NLP), given the limited training data ~\citep{kazemnejad_impact_2023} and compute, models cannot learn from large-scale long sequences and thus are also expected to possess such generalization ability to process long sequences ~\citep{shaham_zeroscrolls_2023}. 
However, it is a challenging task for the de facto Transformer architecture ~\citep{vaswani_attention_2017}, though Transformer-based large language models (LLMs) ~\citep{touvron_llama_2023,openai_gpt-4_2023} have drastically advanced the NLP field. 

Transformer-based models are trained on sequences with a maximum length ~\citep{raffel_exploring_2020,zhang_pegasus_2020,brown_language_2020}, as a result of the quadratic memory and computational complexity with regard to input length. To make matters worse, some research reveals that Transformers might have gained their performance from surface-level memorization instead of abstract, generalizable skills ~\citep{razeghi_impact_2022,wu_reasoning_2024}, which means they can hardly break through the maximum training length and perform poorly on sequences with length beyond it ~\citep{dai_transformer-xl_2019,neishi_relation_2019}, i.e., they cannot perform \textbf{length extrapolation} ~\citep{mitchell_extrapolation_2018, press_train_2021}. To offer a more comprehensive understanding of the challenges in length extrapolation, we present comparison results of three state-of-the-art models with different context sizes on several generation tasks in Appendix \ref{appdix:compa}. 

The length limit together with poor length extrapolation prevents LLMs from handling long sequences, such as DNA and protein sequences ~\citep{abramson_accurate_2024}, high-resolution images ~\citep{liu2023llava}, and even videos ~\citep{lin_video-llava_2023}. Moreover, existing approaches for harnessing the full potential of LLMs also demand a larger context window, to incorporate elaborate prompts ~\citep{liu_pre-train_2023}, sufficient in-context demonstrations ~\citep{brown_language_2020} and long-term memory of agents ~\citep{park_generative_2023}. Hence, there is a growing body of research trying to strengthen length extrapolation of LLMs ~\citep{press_train_2021,ontanon_making_2022,anil_exploring_2022,chi_dissecting_2023,sun_length-extrapolatable_2023}, mostly from the perspective of positional encoding (PE).

\tikzstyle{my-box}=[
    rectangle,
    draw=hidden-draw,
    rounded corners,
    text opacity=1,
    minimum height=1.5em,
    minimum width=40 em,
    inner sep=2pt,
    align=center,
    fill opacity=.5,
    line width=0.8pt,
]
\tikzstyle{leaf}=[my-box, minimum width=1.5em,
    fill=hidden-pink!80, text=black, align=center, text width=40em, font=\normalsize,
    inner xsep=2pt,
    inner ysep=4pt,
    line width=0.8pt,
]

\begin{figure*}[t!]
    \centering
    \resizebox{\textwidth}{!}{
        \begin{forest}
            forked edges,
            for tree={
                grow=east,
                reversed=true,
                anchor=base west,
                parent anchor=east,
                child anchor=west,
                base=center,
                font=\large,
                rectangle,
                draw=hidden-draw,
                rounded corners,
                align=center,
                text centered,
                minimum width=3em,
                edge+={darkgray, line width=1pt},
                s sep=3pt,
                inner xsep=2pt,
                inner ysep=3pt,
                line width=0.8pt,
                ver/.style={rotate=90, child anchor=north, parent anchor=south, anchor=center},
            },
            where level=1{text width=10em,font=\normalsize,}{},
            where level=2{text width=15em,font=\normalsize,}{},
            where level=3{text width=14em,font=\normalsize,}{},
            where level=4{text width=24em,font=\normalsize,}{},
            where level=5{text width=10em,font=\normalsize,}{},
            [\rotatebox{90}{\textbf{Length Extrapolation}}, for tree={fill=a35}
                [\textbf{Extrapolatable PEs} \S\ref{sec:pe}, for tree={fill=paired-orange}
                    [\textbf{APEs}~\citep{vaswani_attention_2017} \S\ref{subsec:abs}, for tree={fill=paired-light-orange}
                        [\textbf{Integrating}\\ \textbf{Shift Invariance} \S\ref{subsubsec:shift}, for tree={fill=paired-dark-yellow}
                        [\textbf{SHAPE} \cite{kiyono_shape_2021}; 
                        \textbf{CAPE} \cite{likhomanenko_cape_2021}
                           , leaf , for tree={fill=paired-dark-yellow}
                        ]]
                        [\textbf{Enhancing Smoothness} \S\ref{subsubsec:smooth}, for tree={fill=paired-dark-yellow}
                        [\textbf{Complex} \cite{wang_encoding_2019};
                        \textbf{FLOATER} \cite{liu_learning_2020} \\ ,leaf, for tree={fill=paired-dark-yellow}]]                  
                    ]
                    [\textbf{RPEs}~\citep{shaw_self-attention_2018} \S\ref{subsec:rel}, for tree={fill=paired-light-orange}
                        [\textbf{RoPE Family} \S\ref{subsubsec:rope}, for tree={fill=paired-dark-yellow}
                        [\textbf{RoPE} \cite{su_roformer_2024};
                        \textbf{xPos} \cite{sun_length-extrapolatable_2023} \\
                        ,leaf, for tree={fill=paired-dark-yellow}]] 
                        [\textbf{T5-bias Family} \S\ref{subsubsec:t5bias}, for tree={fill=paired-dark-yellow}
                        [\textbf{T5-Bias} \cite{raffel_exploring_2020};
                        \textbf{TISA} \cite{wennberg_case_2021};\\
                        \textbf{ALiBi} \cite{press_train_2021};
                        \textbf{KERPLE} \cite{chi_kerple_2022}; \\
                        \textbf{Sandwich} \cite{chi_dissecting_2023};
                        \textbf{FIRE} \cite{li_functional_2023};
                        \textbf{CAPE} \cite{zheng_cape_2024}
                            , leaf, for tree={fill=paired-dark-yellow}
                        ]]]]
                [\textbf{PE-based Methods} \S\ref{sec:methods}, for tree={fill=a26}
                    [\textbf{Position Interpolation} \S\ref{subsec:pi}, for tree={fill=a26}
                        [\textbf{Linear Positional Interpolation} ~\citep{chen_extending_2023};
                        \textbf{NTK-Aware Interpolation} ~\citep{blocntkaware};\\
                        \textbf{Dynamic-NTK Interpolation} ~\citep{emozillareddit};
                        \textbf{NTK-by-parts Interpolation} ~\citep{blocntkparts};
                        \textbf{Truncated Basis} ~\citep{pal_giraffe_2023}
                            , leaf, for tree={fill=paired-cyan, text width=55.6em}
                        ]
                    ]
                    [\textbf{Randomized PE} \S\ref{subsec:randomized}, for tree={fill=a26}
                        [\textbf{Randomized PE} \citep{ruoss_randomized_2023};
                        \textbf{PoSE} \citep{zhu_pose_2023}
                            ,leaf, for tree={fill=paired-cyan, text width=55.6em}
                        ]]]]]
        \end{forest}}
    \caption{Taxonomy for length extrapolation of Transformers.}
    \label{fig:lit_surv}
\end{figure*}
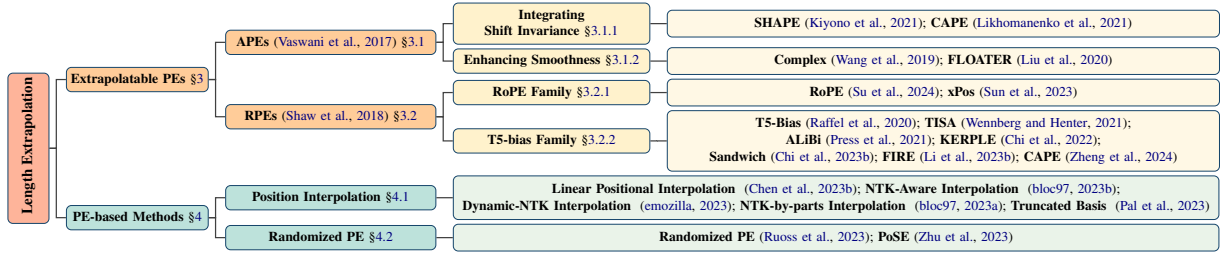
Despite the prosperity in this area, a systematic survey is still lacking. We aim to fill this blank by investigating existing approaches that enable and enhance length extrapolation of Transformers. Specifically, a brief formal introduction to Transformer is given in \S \ref{sec:pre} as a solid foundation for further discussion. Then, we comprehensively summarize extrapolatable PEs proposed from the birth of Transformer to the prevalence of LLMs in \S \ref{sec:pe}. Note that we focus exclusively on PEs proposed for better extrapolation and omit others, since there is already an insightful survey on PEs of Transformer ~\citep{dufter_position_2022}. Based on these PEs, many novel methods emerge in the era of LLMs to further enhance extrapolation, which we intentionally centralize in \S \ref{sec:methods}, covering popular position interpolation methods and randomized methods. These advancements demonstrate the vibrancy and vastness of this area, from which we distill future directions and insights, represented in \S \ref{sec:future} and \S \ref{sec:dis}.

\section{Preliminary}
\label{sec:pre}
In this section, we follow ~\citet{dufter_position_2022} to present a formal description of the encoder layer of the Transformer, as the decoder layer is almost the same except for the cross-attention mechanism.
 Given an input matrix $\boldsymbol X\in\mathbb R^{n\times d}$ as a sequence of $n$ embeddings with dimension $d$, an encoder layer $f:\mathbb R^{n\times d}\xrightarrow{} \mathbb R^{n\times d}$ with $f(\boldsymbol X)=\boldsymbol Z$ is defined by:
  \begin{align}
      \boldsymbol{C} &= \frac{\boldsymbol Q \boldsymbol K^T}{\sqrt d} \label{eq:compa} \\ 
      \boldsymbol{A} &= \text{Softmax}(\boldsymbol C)\boldsymbol V \label{eq:attn} \\ 
      \boldsymbol{O} &= \text{LayerNorm}_1(\boldsymbol A+ \boldsymbol X) \label{eq:attout} \\ 
      \boldsymbol{F} &= \text{ReLU}(\boldsymbol O \boldsymbol W^{(f_1)}+\boldsymbol b^{(f_1)})\boldsymbol W^{(f_2)}+\boldsymbol b^{(f_2)} \label{eq:ffn} \\ 
      \boldsymbol{Z} &= \text{LayerNorm}_2(\boldsymbol O+ \boldsymbol F) \label{eq:ffnout}
  \end{align}
where $\boldsymbol Q=\boldsymbol X \boldsymbol W_q, \boldsymbol K=\boldsymbol X \boldsymbol W_k, \boldsymbol V=\boldsymbol X \boldsymbol W_v$ are queries, keys and values, with $\boldsymbol W_q, \boldsymbol W_k, \boldsymbol W_v\in \mathbb R^{d\times d}$ being the projection matrices. 

Firstly, the compatibility scores $\boldsymbol C$ are computed as the dot product between queries and keys with a scaling factor\footnote{We will omit this scaling factor in the following for simplicity and clarity.} $1/\sqrt{d}$ (Equation \ref{eq:compa}). Then, the row-wise softmax function converts compatibility scores into weights, and the weighted sum of the values is the output of the attention layer (Equation \ref{eq:attn}). The fully connected feed-forward network consists of two linear transformations with a ReLU activation between (Equation \ref{eq:ffn}), with parameters $\boldsymbol W^{(f_1)}\in \mathbb R^{d\times d_f}, \boldsymbol W^{(f_2)}\in \mathbb R^{d_f\times d}, \boldsymbol b^{f(1)}\in \mathbb R^{d_f}, \boldsymbol b^{(f_2)}\in \mathbb R^{d}$, where $d_f$ is the intermediate dimension. Besides, residual connection ~\citep{he_deep_2016} and layer normalization~\citep{ba_layer_2016} are leveraged (Equation \ref{eq:attout} and \ref{eq:ffnout}) to enhance scalability.

Note that in the above descriptions, we have not imposed any limit on input length $n$, which means the Transformer is naturally equipped with a notion of length extrapolation. Theoretically, a fixed setting of Transformer weights defines a sequence-to-sequence function on sequences of \textit{arbitrary length} ~\citep{yun_are_2019}. If the function applies the correct transformation for inputs of any length, it is expected to length extrapolate ~\citep{zhou_what_2023}.

However, we have to break this nature by integrating PE with Transformers to inject position information into them. Otherwise, they are \textbf{permutation equivalent} or \textbf{order invariant}\footnote{Note that some existing research suggests causal language models can learn position information without PE ~\citep{tsai_transformer_2019, haviv_transformer_2022,chi_latent_2023}.}. Thus, PEs are central to length extrapolation and form the core focus of this survey.


\section{Extrapolatable Positional Encodings}
\label{sec:pe}
Sinusoidal position embeddings are proposed with Transformer as it may help extrapolate to longer sequences beyond training~\cite{vaswani_attention_2017}. The idea behind this claim, that length extrapolation can be enabled by simply changing PE, has been widely supported and demonstrated \citep{neishi_relation_2019,press_train_2021,ruoss_randomized_2023}. Hence, developing better PEs has been the predominant avenue to enhance length extrapolation of Transformers. Table \ref{tab:pes} presents a characterization of these extrapolatable PEs.

\begin{figure}
  \centering
  \includegraphics[width=\linewidth]{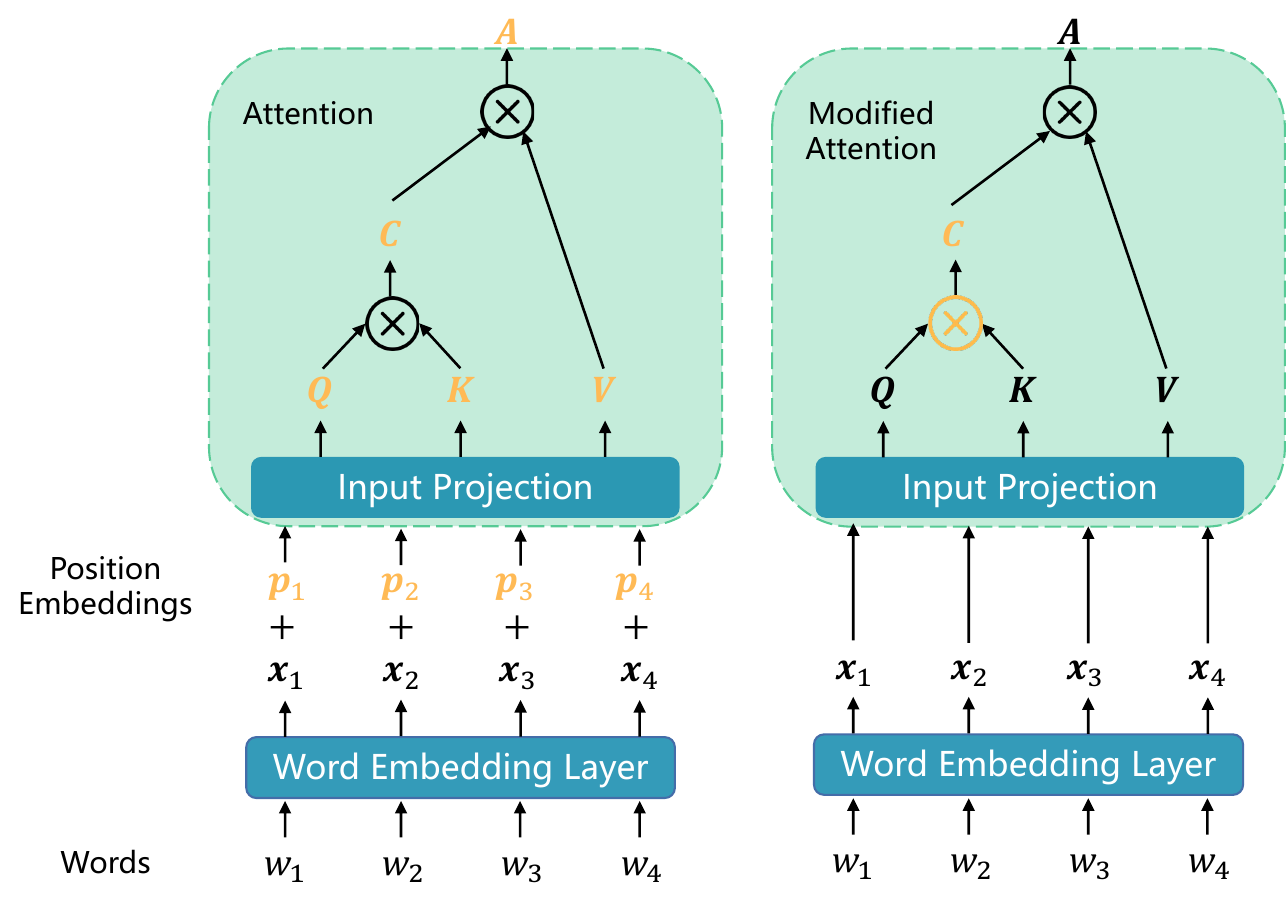}
  \caption{General differences between APE (left part) and RPE (right part), where \textcolor{pic-orange}{orange} denotes elements holding position information.}
  \label{fig:ape&rpe}
\end{figure}

Basically, absolute positional encodings (APEs) map each position to a unique representation and integrate it with corresponding word embedding, while relative positional encodings (RPEs) encode the relative distance between tokens and directly inject it into the attention module.  Besides, RPEs usually keep modifications independent of value vectors and leaves them not entangled with position information. Hence, position information of RPEs can be scalars and usually recurs at each layer. Figure \ref{fig:ape&rpe} illustrates these general differences. We divide Table \ref{tab:pes} and this section based on whether the PE is absolute or relative, as existing research suggests this distinction significantly impacts length extrapolation ~\citep{neishi_relation_2019, likhomanenko_cape_2021,chi_kerple_2022}.

\begin{table}
\centering
\setlength{\tabcolsep}{3pt}
\resizebox{\linewidth}{!}{
\begin{tabular}{ll|cccc}  
\toprule
 &\multirow{2}{*}{PE}& \multirow{2}{*}{Manifestation}& \multirow{2}{*}{Learnable} &\multirow{2}{*}{Integration} & Injection\\
 & & & & & Layer \\
\midrule
\multirow{7}{*}{\rotatebox{90}{APE}}
 &Sinusoidal{\small ~\citep{vaswani_attention_2017}}   & Embedding  & \ding{53}      &Add     &Initial\\
 \cmidrule(r){2-6} 
 & \small{\textit{with Shift Invariance}} \\
 &SHAPE{\small ~\citep{kiyono_shape_2021}}             & Embedding  & \ding{53}      &Add     &Initial\\
 &CAPE{\small ~\citep{likhomanenko_cape_2021} }        & Embedding  & \ding{53}      &Add     &Initial\\
 \cmidrule(r){2-6} 
 & \small{\textit{with Smoothness}} \\
 &Complex{\small ~\citep{wang_position_2020}}          & Embedding  & \ding{51}   &Multiply     &Initial\\
 &FLOATER{\small ~\citep{liu_learning_2020} }          & Embedding  & \ding{51}   &Add     &Initial\\
 \hline
 \multirow{11}{*}{\rotatebox{90}{RPE}}
 &{\small ~\citet{shaw_self-attention_2018} }              & Embedding & \ding{51}     &Add  &Every \\
 \cmidrule(r){2-6} 
 & \small{\textit{T5 Family}} \\
 &T5 Bias{\small ~\citep{raffel_exploring_2020} }          & Bias      & \ding{51}     &Add  &Every \\
 &\textbf{ALiBi}{\small ~\citep{press_train_2021}}         & Bias      & \ding{53}        &Add  &Every \\
 &\textbf{KERPLE}{\small ~\citep{chi_kerple_2022} }        & Bias      & \ding{51}     &Add  &Every \\
 &\textbf{SANDWICH}{\small ~\citep{chi_dissecting_2023}}   & Embedding & \ding{53}        &Add  &Every \\
 &\textbf{FIRE}{\small ~\citep{li_functional_2023} }       & Bias      & \ding{51}     &Add  &Every \\
 &\textbf{CAPE}{\small ~\citep{zheng_cape_2024} }       & Bias      & \ding{51}     &Add  &Every \\
 \cmidrule(r){2-6} 
 & \small{\textit{RoPE Family}} \\
 &\textbf{RoPE}{\small ~\citep{su_roformer_2024} }         & Embedding & \ding{53}        &Multiply  &Every \\
 &\textbf{xPOS}{\small ~\citep{sun_length-extrapolatable_2023} }         & Embedding & \ding{53}        &Multiply  &Every \\
\bottomrule
\end{tabular}}
\caption{A list of extrapolatable PEs. \textbf{Bolded} methods are proposed or widely adopted for LLMs. \textit{Manifestation} shows how the position infomation is introduced. \textit{Learnable} shows whether it can adjust based on the input. \textit{Integration} shows how the position representations are integrated with token representations. \textit{Injection Layer} shows the injecting position PE.}
\label{tab:pes}
\end{table}

\subsection{Absolute Positional Encodings}
\label{subsec:abs}
Specifically, for a token in position $pos$, the sinusoidal position embedding is defined as:
\begin{equation}
    [\dots,\sin(\frac{pos}{10000^{2i/d}}),\cos(\frac{pos}{10000^{2i/d}}),\dots],
\end{equation}
where $i\in [0,d/2-1]$ is the dimension of the position embedding and $d$ denotes model dimension. Then, each position embedding is added to the corresponding token embedding and the sum is fed into Transformer, so the compatibility score between query $\boldsymbol q_i$ and key $\boldsymbol k_j$ can be formalized as
\begin{align}
    \boldsymbol q_i\boldsymbol k_j^T&=((\boldsymbol x_i+\boldsymbol p_i)\boldsymbol W_q)((\boldsymbol x_j+\boldsymbol p_j)\boldsymbol W_k)^T. \label{eq:abs}
\end{align}
This equation is the basis of many other PEs. 

However, researchers subsequently found that sinusoidal APE is hard to extrapolate ~\citep{dai_transformer-xl_2019,neishi_relation_2019}. Hence, a wide variety of APEs have been proposed to enhance sinusoidal APE and extrapolation of Transformers from different perspectives, either trying to integrate shift invariance in sinusoidal APE (\S \ref{subsubsec:shift}) or aiming to generate position embeddings varying smoothly with position indices (\S \ref{subsubsec:smooth}).

\subsubsection{Integrating Shift Invariance}
\label{subsubsec:shift}
Taking inspiration from the three properties of PEs proposed by ~\citet{wang_position_2020}, ~\citet{kiyono_shape_2021} speculated superior extrapolation performance comes from \textit{shift invariance}, the property of a function to not change its output even if its input is shifted. Aiming to incorporate the benefit of shift invariance in sinusoidal APE, they simply shift every position index of a sequence by a random offset $k$ during training, which prevents the model from using absolute positions and instead encourages the use of relative positions. 

Following a similar idea, ~\citet{likhomanenko_cape_2021} took it a step further by leveraging continuous signals. In addition to shifting every position index of APE by an identical random offset, which they call \textit{global shift}, they also introduced \textit{local shift}, i.e., shifting each position index by a different random shift, and \textit{global scaling}, i.e., scaling every position index by an identical random scalar, to further prevent capturing spontaneous correlations and memorizing distances.

\subsubsection{Enhancing Smoothness}
\label{subsubsec:smooth}
Apart from above relatively straightforward methods based on sinusoidal APE, there are several APEs taking quite different theoretical avenues to enhance length extrapolation, aiming to improve the smoothness of the position representations.

~\citet{wang_encoding_2019} proposed to extend each word embedding as a continuous function over an independent variable, i.e., position, so that word representations vary smoothly with increasing positions. Through mathematically sound derivation, their general complex-valued embedding $f(j,pos)$ of a word $w_j$ in position $pos$ is
\begin{equation}
    [r_{j,1}e^{i(\omega_{j,1}pos+\theta_{j,1})},\cdots,  r_{j,d}e^{i(\omega_{j,d}pos+\theta_{j, d})}],
\end{equation}
where amplitude $\boldsymbol r=[r_{j,1},\dots,r_{j,d}]$, frequency $\boldsymbol \omega=[\omega_{j,1},\dots, \omega_{j,d}]$ and initial phrase $\boldsymbol \theta=[\theta_{j,1},\dots,\theta_{j,d}]$ are all trainable. In addition to representing positions in complex plane for the first time, multiplying position embeddings with word embeddings is another of their innovations.

An alternative approach is to directly capture the dynamics between position representations.  ~\citet{liu_learning_2020} introduced a dynamical system to model position representations $\{\boldsymbol p_i\in \mathbb R^d:i=1,\dots, n\}$, which can be characterized as
\begin{equation}
\resizebox{.95\linewidth}{!}{$
\displaystyle
    \boldsymbol p(t)=\boldsymbol p(s)+\int_s^t \boldsymbol{h}(\tau, \boldsymbol{p}(\tau);\boldsymbol{\theta}_h)d\tau,0\le s\le t< \infty
    $}
\end{equation}
with an initial vector $\boldsymbol p(0)$, where $\boldsymbol p(t):\mathbb R_+\mapsto\mathbb R^d$ is the continuous version of the discrete sequences $\{\boldsymbol p_i\}$. $\boldsymbol{h}(\tau, \boldsymbol{p}(\tau);\boldsymbol{\theta}_h)$, which is the "latent force" that drives the changes from $\boldsymbol p_i$ to $\boldsymbol p_{i+1}$, is actually a neural network parameterized by $\boldsymbol \theta_h$ and takes in the previous state $(\tau,\boldsymbol p(\tau))$.

\textbf{Highlights:} As the first PE for Transformer, sinusoidal APE has a significant impact on PEs thereafter, despite its poor extrapolation. To improve this, researchers either leverage random shift to incorporate shift invariance in sinusoidal APE or generate position embeddings varying smoothly with position. Among them, simple random shifting is like a small patch for sinusoidal APE and has limited benefits for extrapolation, at the cost of possible semantic confusion in position encoding, while the latter can hopefully lead to better extrapolation, coming with a much higher parameter- and computation-complexity.

\subsection{Relative Positional Encodings}
\label{subsec:rel}
Albeit for the efforts in extrapolatable APEs, it is believed that RPEs are theoretically capable of running on unseen lengths and are more robust to input length change ~\citep{neishi_relation_2019,likhomanenko_cape_2021,chi_kerple_2022}, as RPEs only rely on relative position information, which means they encode the idea of shift invariance naturally and are not subject to a maximum position value. Besides, there is a consensus that in natural language, it is not absolute but relative position that matters ~\citep{huang_improve_2020, sinha_curious_2022}. Thus, RPEs become the dominant way to encode positions, which we detail in this section. Before that, we reformulate Equation \ref{eq:abs} as follows to clarify the perspective of RPEs:
\begin{equation}
    \boldsymbol q_i\boldsymbol k_j^T=(\boldsymbol x_i\boldsymbol W_q )(\boldsymbol x_j\boldsymbol W_k)^T\oplus\boldsymbol p(j-i), \label{eq:rel}
\end{equation}
where $\boldsymbol p(j-i)$ encodes the relative position information, $\oplus$ denotes any approach of integrating the position information into the compatibility score.

Among the first, ~\citet{shaw_self-attention_2018} introduced the idea of RPE based on above formulation. Specifically, they concretized Equation \ref{eq:rel} as
\begin{equation}
    \boldsymbol q_i\boldsymbol k_j^T=(\boldsymbol x_i\boldsymbol W_q)(\boldsymbol x_j\boldsymbol W_k+\boldsymbol p_r)^T, \label{eq:shaw}
\end{equation}
where $\boldsymbol p_r\in \mathbb R^d$ is a trainable relative position embedding and $r=\text{clip}(j-i, r_{\min},r_{\max})$ denotes the clipped relative position. By clipping the relative positions to a determined range, the number of position embeddings to be learned is reduced and length extrapolation is enhanced as unseen position embeddings are avoided. This RPE can also be regarded as a derivation of sinusoidal APE. Following this line, more RPEs have been proposed to better model position information, such as ~\citet{dai_transformer-xl_2019}, ~\citet{huang_improve_2020} and TUPE ~\citep{ke_rethinking_2020}. We omit them here since they are not proposed for stronger length extrapolation.

\subsubsection{RoPE Family}
\label{subsubsec:rope}
Also inspired by sinusoidal APE, ~\citet{su_roformer_2024} proposed to multiply keys and queries by rotation matrices, leaving compatibility scores as
\begin{align}
    \boldsymbol q_i^T\boldsymbol k_j&=(\boldsymbol R_{\Theta,i}^d\boldsymbol x_i\boldsymbol W_q)^T(\boldsymbol R_{\Theta,j}^d\boldsymbol x_j\boldsymbol W_k)\nonumber \\
    &=\boldsymbol W_q^T\boldsymbol x_i^T\boldsymbol R_{\Theta,j-i}^d\boldsymbol x_j \boldsymbol W_k,
\end{align}
where $\boldsymbol R_{\Theta,j-i}^d=(\boldsymbol R_{\Theta, i}^d)^T\boldsymbol R_{\Theta,j}^d$ with $\boldsymbol R_{\Theta, i}^d$ being a block-diagonal matrix with rotation matrices
\begin{equation}
    \begin{pmatrix}
        \cos i\theta_m & -\sin i\theta_m\\
        \sin i\theta_m & \cos i\theta_m
    \end{pmatrix}
\end{equation}
on its diagonal, given the parameters $\Theta=(\theta_m)_{m=1,2,\dots,d/2}$ where $\theta_m=10000^{-2(m-1)/d}$. Here the \textit{base} is $10000$, and $\lambda_m=2\pi/\theta_m$ is wavelength. This method is called Rotary Position Embedding (RoPE) as intuitively it rotates key/value embeddings according to their position index:
\begin{equation}
f_{\{q,k\}}(\boldsymbol x_i,i)=\boldsymbol R_{\Theta,i}^d\boldsymbol x_i \boldsymbol W_{\{q,k\}}\label{eq:rope}.
\end{equation} 
It is noteworthy that despite the absolute nature of this rotary process, the compatibility score and thus attention depend only on relative distance. This property together with long-term decay for inter-token product benefit length extrapolation.

As RoPE has been widely used in popular LLMs ~\citep{touvron_llama_2023,jiang_mistral_2023,anil_palm_2023}, there are some variants proposed to improve it. \citet{sun_length-extrapolatable_2023} defined attention score expectation between two tokens at a specific distance and further attributed the poor extrapolation of RoPE to the dramatic oscillation of their attention expectations. They proposed to fix this issue by incorporating a balancing term to punish the oscillation of unstable dimensions and keep the distribution of stable ones, which can be simplified to:
\begin{equation}
    \boldsymbol q_i^T\boldsymbol k_j=\gamma^{i-j}\boldsymbol W_q^T\boldsymbol x_i^T\boldsymbol R_{\Theta,j-i}^d\boldsymbol x_j \boldsymbol W_k,
\end{equation}
where $\gamma\in (0,1)$ is a scalar hyperparameter.

\subsubsection{T5-Bias Family}
\label{subsubsec:t5bias}
Different from complex embedding form, some researchers reduce position information $\boldsymbol{p}(j-i)$ to a simpler form. ~\citet{raffel_exploring_2020} utilized learnable scalars to represent relative position information:
\begin{equation}
    \boldsymbol q_i \boldsymbol k_j^T=(\boldsymbol x_i\boldsymbol W_q)(\boldsymbol x_j \boldsymbol W_k)^T+\beta_{i,j}. \label{eq:t5}
\end{equation}
In addition, they extended the clipping mechanism by a logarithmic bucket assignment to achieve precise discrimination of nearby positions and less precise discrimination of further positions (e.g., mapping the position indices 1-4 to themselves, 5-6 to 5, 7-8 to 6, 9-12 to 7, and so forth.), which further reduces the parameters to be learned and is beneficial for extrapolation ~\citep{chi_kerple_2022}. Moreover, \citet{wennberg_case_2021} introduced TISE, which leverages a radial-basis function of relative distance with multiple trainable parameters to add a bias to attention scores.


As the first PE aiming mainly for length extrapolation, ALiBi ~\citep{press_train_2021} takes an even simpler way to represent relative position:
\begin{equation}
    \boldsymbol q_i \boldsymbol k_j^T=(\boldsymbol x_i\boldsymbol W_q)(\boldsymbol x_j\boldsymbol W_k)^T+m(j-i),
\end{equation}
where scalar $m$ is a head-specific slope fixed before training. It is worth noting that there is no additional learnable parameter, which leads to superior efficiency and may also contribute to better extrapolation of ALiBi. Empirical experiments on language modeling demonstrated its superiority.


From the perspective of kernel methods, ~\citet{chi_kerple_2022} considered ALiBi as a triangle kernel and extended it to KERPLE, a framework that generalizes relative position embeddings for extrapolation by kernelizing positional differences using conditionally positive definite kernels. In this framework, various RPEs can be derived from different conditionally positive definite kernels in a principled way, among which the logarithmic variant achieves preferred extrapolation performance, by calculating the compatibility score as follows:
\begin{equation}
    \boldsymbol q_i^T\boldsymbol k_j=(\boldsymbol x_i \boldsymbol W_q)^T(\boldsymbol x_j\boldsymbol W_k)-r_1\cdot\log(1+r_2|i-j|),
\end{equation}
where $r_1,r_2$ are positive scalar parameters.

Aware of the overfitting issue of sinusoidal APE, ~\citet{chi_dissecting_2023} proposed to overcome it by simplifying sinusoidal APE to a new RPE, Sandwich. Specifically, they dropped the cross terms in Equation \ref{eq:abs} and kept the inner product of two position embeddings as position information:
\begin{equation}
    \boldsymbol q_i^T \boldsymbol k_j=(\boldsymbol x_i\boldsymbol w_q)^T(\boldsymbol x_j\boldsymbol W_k)+\boldsymbol p_i^T\boldsymbol p_j.
\end{equation}
It is worth noting that in this formula, $\boldsymbol p_i^T\boldsymbol p_j$ becomes a temporal bias term with the same decay-with-distance pattern as ALiBi, which is exactly what the authors want to achieve as they suggested this pattern is likely to be the secret to successful length extrapolation. Besides, since position embeddings here only need to interact with themselves, the authors make the dimension of them a hyperparameter to further improve performance.

FIRE ~\citep{li_functional_2023} integrates positional information into Transformers following T5 bias:
\begin{equation}
    \boldsymbol q_i \boldsymbol k_j^T=(\boldsymbol x_i \boldsymbol W_q)(\boldsymbol x_j \boldsymbol W_k)^T+b(i,j),
\end{equation}
where the bias $b(i,j)$ is mapped from positions using a learnable continuous function $f_\theta:\mathbb R\xrightarrow{} \mathbb R$, e.g., MLP. To avoid the generalization issue when the inputs are outside the training domain of the function, they proposed progressive interpolation by normalizing the distance by query position index, namely $b(i,j)=f_\theta(\frac{i-j}i)$. Note that in causal attention, the normalized distance is always bounded between $[0,1]$, which aligns the inference domain with the training domain for any sequence lengths, leading to better length extrapolation.

However, the above methods separate positional bias from semantics completely, which may cause semantic similarity to be overshadowed by position information. Hence, ~\citet{zheng_cape_2024} proposed Context-Adaptive Positional Encoding (CAPE) to integrate both semantic and positional information:
\begin{align}
    \boldsymbol{q}_i\boldsymbol{k_j}^T&=(\boldsymbol x_i \boldsymbol W_q)(\boldsymbol x_j \boldsymbol W_k)^T \nonumber \\
    &+f((\boldsymbol x_i \boldsymbol W_q)(\boldsymbol x_j \boldsymbol W_k)^T,b(i,j)).
\end{align}
Here $f:\mathbb R\times \mathbb R\to \mathbb R$ is parameterized by a two-layer LeakyReLU neural network and $b(i,j)$ come from other RPEs(e.g., ALiBi and FIRE).

In addition to RPEs introduced previously, there are some methods cannot be categorized into RoPE or T5-bias family. ~\citet{he_two_2024} introduce bilevel PE that employs two distinct PE for each position: an APE for intra-segment position to help model capture the semantics contained therein, while an RPE for inter-segment position to capture relationships between segments and exhibits extrapolation. This decoupling offers greater flexibility in addressing the length extrapolation problem. 

Based on the observation that existing PEs use token as the unit of measurement, ~\citet{golovneva_contextual_2024} claimed that this feature prevents PEs from generalizing to higher levels of abstraction such as sentences and paragraphs. Therefore, they proposed Contextual Positional Encoding (CoPE), which allows the model to determine semantic unit (e.g., word and sentence) and assign tokens therein a same position index. Since CoPE can distribute positions to a much larger number of tokens and focus attention on semantic units at a higher level of abstraction, it exhibits stronger extrapolation.

\textbf{Highlights:} Earlier RPEs had been greatly influenced by sinusoidal APEs by modifying terms in Equation \ref{eq:abs} and replacing absolute embeddings with relative embeddings. These methods usually leverage clipping or binning strategy to avoid out-of-distribution position embeddings and enhance extrapolation. Since RPEs decouple the one-to-one correspondence between position and position representation, incorporating bias term directly into compatibility score (Equation \ref{eq:rel}) becomes a feasible and even better way to encode positional information, which is much simpler and naturally disentangles value vectors and position information. However, despite the strong extrapolation of these bias methods, they cannot represent complex distance-attention functions based on Fourier basis like RoPE. Therefore, RoPE become the de facto PE of recent LLMs due to its advanced general performance, in spite of its poor extrapolation. 

\section{Extrapolation Methods in LLMs Era}
\label{sec:methods}
\begin{figure}
  \centering
  \includegraphics[width=\linewidth]{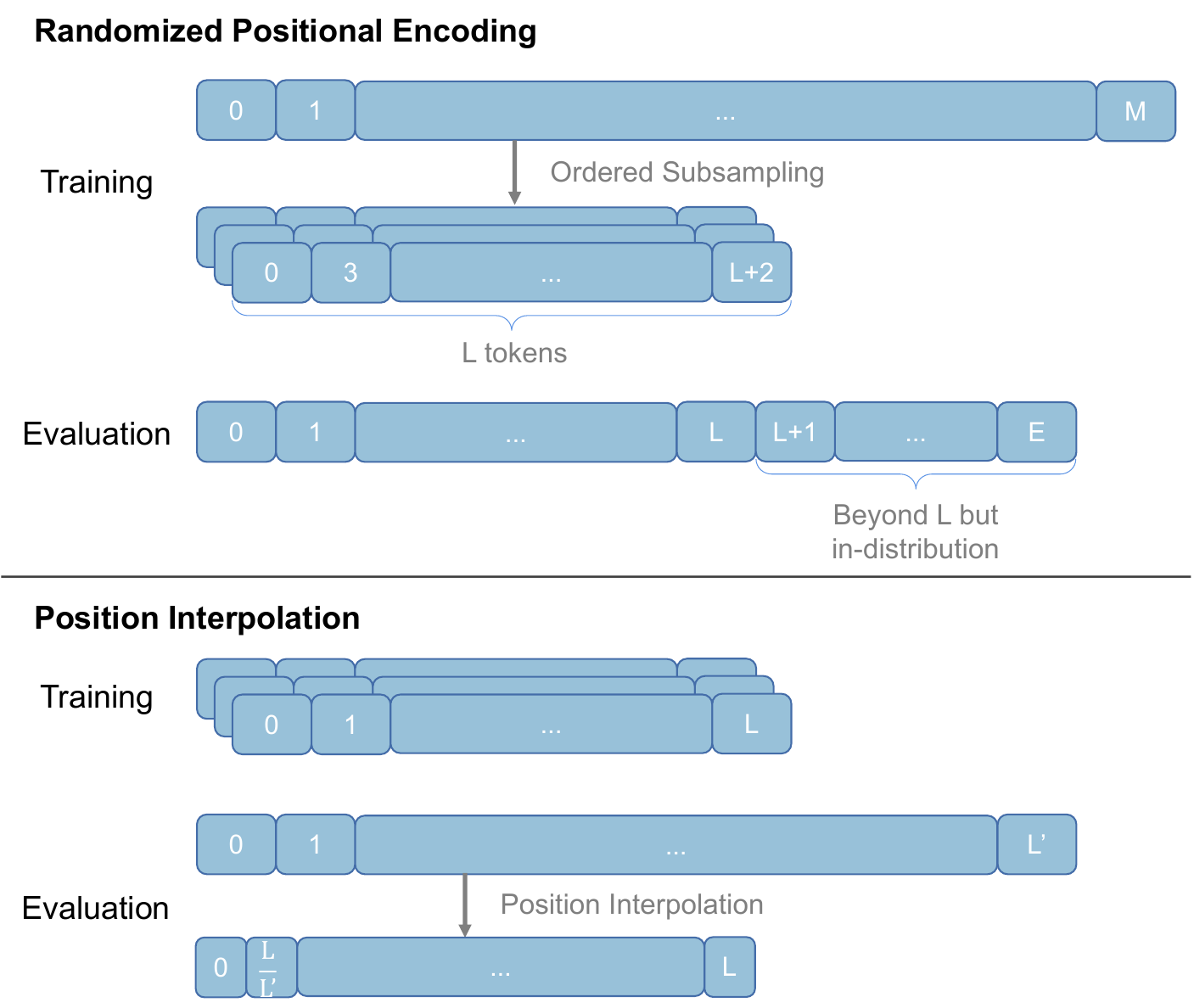}
  \caption{Essentials of position interpolation and randomized PE. Randomized PE aims to ensure that positions falling outside the context window at inference remain in distribution through advanced exposure in training. Position interpolation, on the other hand, works during the inference stage by scaling a longer position range into the original context window.}
  \label{fig:methods}
\end{figure}

Based on PEs in \S \ref{sec:pe}, various methods have been developed to further enhance length extrapolation of LLMs. This section is separated in response to this wave, focusing on interpolation methods and randomized PEs, as illustrated in Figure \ref{fig:methods}.

\subsection{Position Interpolation}
\label{subsec:pi}
Despite the large quantity of PEs with better extrapolation, RoPE has been most widely adopted in recent LLMs due to its superior in-distribution performance. Hence, loads of methods have been proposed to enhance the extrapolation of RoPE, the most prevalent of which is position interpolation.

~\citet{chen_extending_2023} firstly \footnote{There is a concurrent work: \url{https://kaiokendev.github.io/til\#extending-context-to-8k}} introduced position interpolation for RoPE to extrapolate LLMs to longer sequences by applying linear scaling to down-scale position indices so that the maximum position index matches the previous length limit during pre-training. Formally, this method replaces RoPE $f$ (Equation \ref{eq:rope}) by $f'$ defined as $f'(\boldsymbol x,i) = f(\boldsymbol x, \frac{iL}{L'})$, where $L$ is the length limit during pre-training and $L'$ is the longer sequence length at inference. The \textit{scale ratio} $\kappa=L'/L$ transforms position $n$ to $n/\kappa$. This method reduces absolute position indices from $[0,L')$ to $[0,L)$ and maximum relative distance from $L'$ to $L$, aligning the ranges of position indices and relative distances to mitigate effects on attention score computation.

However, from the perspective of Neural Tangent Kernel (NTK) theory ~\citep{jacot_neural_2018}, simply interpolating RoPE's Fourier space linearly will cause the loss of high-frequency information and prevent models from distinguishing nearby positions.Hence, NTK-Aware Scaled RoPE (NTK-aware interpolation)~\citep{blocntkaware} has been proposed by modifying the base of RoPE:
\begin{equation}
    \theta_m^* = (b\cdot \kappa^{\frac{d}{d-2}})^{-2(m-1)/d},
\end{equation}
where $b$ is the original base and $\kappa$ is still the scale ratio. The core idea here is to scale high frequencies less and low frequencies more to reduce information loss of high frequencies. As NTK-aware interpolation does not scale the Fourier features directly, all positions are distinguishable from each other. Moreover, this method does not require any fine-tuning to extend the context window.

Further, Dynamic-NTK interpolation~\citep{emozillareddit} combined NTK-aware interpolation with dynamic scaling, using exact positions for tokens within pre-trained context window to prevent performance degradation and dynamically increases scale ratio $\kappa$ as current sequence length increases to adjust positions beyond the window:
\begin{equation}
    \kappa=\begin{cases}
        L'/L,&\text{if}\ L'/L>1, \\
        1,&\text{otherwise},
    \end{cases}
\end{equation}
where $L'$ is the sequence length of the current sequence, which will increase after each step.

Either scaling position indices or modifying bases, all position representations become closer to each other, impairing LLM's ability to distinguish the positional order of close-by tokens. Besides, ~\citet{blocntkparts} observed that some RoPE dimensions have wavelengths longer than the pre-trained context window, where they presume absolute positional information remains intact\footnote{From the perspective of frequency, the full range of high-frequency components have been seen by the model during training, while low-frequency components have not. Thus, every position within the context window leads to a unique value in these low-frequency components, based on which models can determine the absolute position of each token.}. Hence, they proposed NTK-by-parts, which does not interpolate dimensions of small wavelengths at all while always interpolating those of big ones. 

Similar observations with NTK-by-parts have been made by ~\citet{pal_giraffe_2023}, based on which they proposed to use the truncated basis:
\begin{equation}
    \theta_i^*=\begin{cases}
        \theta_i& \text{for}\ \theta_i\ge b,\\
        \rho& \text{for}\ a<\theta_i<b,\\
        0& \text{for}\ \theta_i<a.
    \end{cases}
\end{equation}
where $\rho$ is a fixed value that is relatively small, and $a$ and $b$ are chosen cutoff values. This way, models will experience all values of the basis in the context length used during fine-tuning by choosing appropriate cutoff values, and are supposed to extrapolate better during inference.

Additionally, ~\citet{peng_yarn_2023} observed that by introducing a temperature $t$ into compatibility score before Softmax, perplexity decreases consistently. 
Combining this finding with NTK-by-parts interpolation, they subsequently proposed YaRN that surpasses previous interpolation methods in both fine-tuned and non-fine-tuned scenarios.

The interpolation methods reflect the critical impact of the rotary base of RoPE on length extrapolation, prompting efforts to enhance extrapolation of RoPE-based LLM by fine-tuning it with a scaled base ~\citep{xiong_effective_2023,roziere_code_2023,liu_scaling_2023}. However, fixed scaling factors overlook the gradual length-extension process and impair performance at shorter lengths, leading to the proposal of dynamic scaling methods ~\citep{chen_clex_2023,zhang_extending_2024,ding_longrope_2024}. Innovatively, ~\citet{wang_resonance_2024} scale each dimension's base by rounding its wavelength to the nearest integer, avoiding phase shifts after each full rotation.

\textbf{Highlights:} Recently, position interpolation methods have raised widespread interest in the research community, as a natural result of their superior extrapolation performance and extremely low overhead. Current interpolation methods either interpolate position indices or RoPE's base, guided by sound theoretical intuition. Besides, different from other extrapolation methods, position interpolation methods have already seen their presence in the open-source models~\citep{bai_qwen_2023,touvron_llama_2023-1,ai_yi_2024}.

\subsection{Randomized Positional Encoding}
\label{subsec:randomized}
For PEs without clipping mechanism, length extrapolation means positions beyond those that have been observed during training, leading to out-of-distribution position representations and thus performance degradation. To address this, an intuitive way is enabling models to observe all possible position representations during training, which is exactly the core idea behind randomized PEs.

As a realization of this idea, ~\citet{ruoss_randomized_2023} proposed to simulate a much longer range of positions ($M$) and randomly selects an ordered subset to fit the training context window for each iteration. Thus, through adequate training, we can ensure that the model encounters enough unique positions and all $M$ positions have been fully trained, leading to consistent extrapolation performance.

Different from ~\citet{ruoss_randomized_2023}, PoSE ~\citep{zhu_pose_2023} partitions a sequence into chunks and adjusts the position indices by adding distinct skipping bias terms between chunks. Hence, PoSE keeps the positions continuous in each chunk, which bears a close resemblance to pre-training, while simultaneously help the model adapt to all positions within a longer context window.

\textbf{Highlights:} Essentially, randomized PEs simply decouple the trained context window with the longer inference one by introducing randomized positions during training or fine-tuning, boosting exposure of all possible positions in advance. This idea is quite different from that of position interpolation methods, where the latter tries to interpolate positions during inference to make them fall into the trained range. For the same reason, position interpolation methods are mostly plug-and-play while randomized PEs usually need further fine-tuning, which makes position interpolation much more appealing due to its low overhead.

\section{Future Directions}
\label{sec:future}
\textbf{Evaluation and Benchmark.} Initially, researchers evaluated length extrapolation by training models on sequences with a length limit and testing them on slightly longer sequences ~\citep{liu_learning_2020, likhomanenko_cape_2021}. During this phase, evaluation samples and metrics came from various downstream tasks such as machine translation and question answering. Given the demonstrated versatility of pre-trained language models in various downstream tasks ~\citep{raffel_exploring_2020, brown_language_2020}, language modeling and perplexity have emerged as the standard metrics for evaluating length extrapolation ~\citep{press_train_2021, haviv_transformer_2022}. Thus, we statistically present some empirical results of trending PEs on language modeling in Appendix \ref{appdix:lm}. However, it has become clear that perplexity alone does not adequately reflect downstream task performance and is insufficient ~\citep{tay_scale_2021, kazemnejad_impact_2023, pal_giraffe_2023, hu_can_2024}. Therefore, dedicated benchmarks and evaluation methods are needed to further advance the field of length extrapolation.

To stimulate subsequent research, we present several preliminary thoughts on the construction of a standardized benchmark in Appendix \ref{appdix:benchmark}.

\textbf{Explainability and Principle.} Despite the remarkable progress, our understanding of length extrapolation remains limited, lacking a general and solid theoretical foundation. The decaying-with-distance pattern was initially thought to be crucial for extrapolatable PEs~\citep{press_train_2021, su_roformer_2024}, but it was later shown to merely accommodate the recency bias of language modeling ~\citep{chi_attention_2023}. Although ~\citet{qinExploringTransformerExtrapolation2024} further provided a theoretical analysis and elaborated that exponential convergence is a sufficient condition for RPEs to length extrapolate, their definition of length extrapolation is also based on language modeling and perplexity, which may limit the applicability of their theorem. Besides, extrapolation methods tend to \emph{avoid} out-of-distribution positions via interpolation or advanced exposure. Thus, it remains unclear when or if Transformers length extrapolate in real-world scenarios and whether or how existing methods help with it.

\textbf{Long Context Utilization.} Existing length extrapolation methods mostly focus on expanding context window of Transformers, while much less attention has been paid to the investigation and optimization of the utilization of long context. In fact, as a result of recent advances, state-of-the-art LLMs are claimed to be capable of processing sequences with up to 128k tokens ~\citep{abdinPhi3TechnicalReport2024, aiMistralNeMo2024}. Given such a long context, the extent to which the models can effectively utilize it becomes a critical question. Previous study has revealed that LLMs tend to "lost in the middle" ~\citep{liu_lost_2023}, i.e., they cannot effectively leverage information in the middle of a long context. Despite a few preliminary explorations trying to improve long context utilization ~\citep{staniszewski_structured_2023,ravautContextUtilizationSummarization2024}, recent long-context benchmarks \citep{liHowLongCan2023,anLEvalInstitutingStandardized2024,baiLongBenchBilingualMultitask2024,zhangInftyBenchExtendingLong2024} suggest that trending long-context LLMs still struggle on long sequences, and significant advancements are required.

\section{Discussions}
\label{sec:dis}
\subsection{Length-Extrapolated and Long-Context Transformers}
Throughout this survey, we position length extrapolation as a promising avenue towards long-context transformers. However, as stated in \S \ref{sec:intro}, it's the length limit and poor length extrapolation together that prevents transformers from processing long sequences, thus the more direct way to extend the context window is to simply relax the length limit.

The most intuitive way to achieve large context window is directly pre-training the model or fine-tuning (continual pre-training) a pre-trained model on long sequences. \citet{xiong_effective_2023} empirically demonstrated that long context continual pre-training is more efficient and similarly effective compared to pre-training from scratch with long sequences. However, both pre-training and fine-tuning (continual pre-training) are costly and demand large-scale high-quality long data, which is scarce ~\citep{kazemnejad_impact_2023}. To reduce memory and computational overhead during training, recurrent Transformer variances integrate recurrence with attention ~\citep{dai_transformer-xl_2019,bulatov_recurrent_2022} while efficient Transformer variants ~\citep{tay_efficient_2022, fournier_practical_2023} mainly aim at improving the quadratic complexity of attention mechanism, but both usually compromise some of the modeling capability and still need large-scale long sequence data. Flash Attention ~\citep{dao_flashattention_2022,dao_flashattention-2_2023} greatly improves both training and inference efficiency of Transformers with little to no overhead, leading to models with much larger context window ~\citep{jiang_mistral_2023,gunasekarTextbooksAreAll2023,liHowLongCan2023}.

On the other side, there are more radical research efforts that attempt to abandon attention and its quadratic complexity with regard to sequence length completely, such as S4 ~\citep{gu_efficiently_2022}, RWKV ~\citep{pengRWKVReinventingRNNs2023}, and Hyena ~\citep{poli_hyena_2023}. 
Further, some recent studies have attempted to scale these novel architectures to billions of parameters, leading to the emergence of Mamba ~\citep{gu_mamba_2023} and RWKV-5/6 ~\citep{pengEagleFinchRWKV2024}. However, it has been demonstrated that Transformer models perform dramatically better than state space models like S4 at copying and retrieving information from context ~\citep{jelassi_repeat_2024}. Thus, whether these novel architectures are better than Transformer and how they perform on real-world scenarios remains to be evaluated.

\subsection{Length Extrapolation and Generalization}
In parallel to research efforts that deem length extrapolation as a promising approach to extend context window of LLMs, another line of research treats it as a generalization problem and analyzes the length generalization behavior of Transformers within small context window on synthetic tasks such as arithmetic and deductive reasoning in a controlled setup ~\citep{lake_generalization_2018,dubois_location_2020,abbe_generalization_2023}, where some intriguing observations and insights have been discovered.

One common observation is that Transformers often struggle with length generalization, whether they are trained from scratch on synthetic tasks ~\citep{lee_teaching_2023, kazemnejad_impact_2023}, fine-tuned from pre-trained LLMs ~\citep{anil_exploring_2022} or tested in in-context learning ~\citep{saparov_testing_2023}. 

As explanations, ~\citet{dziri_faith_2023} hypothesize certain tasks may not possess the inherent compositionality and allow for shortcut pattern matching. On the other side, Transformers are proven to length generalize on specific tasks ~\citep{zhou_what_2023, xiao_theory_2024} or with the right combination of data format and PE ~\citep{zhou_transformers_2024}. Meanwhile, some studies show other factors in length generalization. ~\citet{anil_exploring_2022} find that fine-tuning regime, scaling data, model sizes, and compute does not improve length generalization, while scratchpad ~\citep{nye_show_2022} or chain-of-thought ~\citep{wei_chain--thought_2022} in the in-context learning regime do. In addition, ~\citet{kazemnejad_impact_2023} show that explicit PE is not essential for decoder-only Transformer to length generalize on small-scale synthetic tasks. These studies have deepened our understanding of length extrapolation in a mechanistic way and broadened our perspectives to go beyond PE, demonstrating that the extrapolation ability needs a systematic design where PE is crucial but by no means the sole component.

\section{Conclusion}
Through this survey, we systematically summarized existing methods and recent advances in length extrapolation from the perspective of PE. Specifically, we meticulously categorize extrapolatable PEs and further dive into methods based on these PEs in LLMs era. In addition, we highlight existing challenges and identify new trends in this research field, hoping to facilitate researchers and provide stimuli for future research.


\section*{Limitation}
This survey presented a systematic review of existing methods and recent trends in length extrapolation of Transformers. However, due to the lack of standardized benchmark and evaluation methods, we primarily focus on high-level comparisons and distinctions in principle of different approaches, rather than fine-grained 
empirical analysis. Furthermore, in this work, we focus on length extrapolation studies aimed at extending the context window of LLMs in real-world scenarios. Although we acknowledge the importance of studies analyzing length generalization in synthetic tasks within a small context window as well, we provide only a brief discussion on them due to the page limitation.

\section*{Acknowledgements}
Xiaocheng Feng is the corresponding author of this work, We thank the anonymous reviewers for their insightful comments. This work was supported by the National Natural Science Foundation of China (NSFC) (U22B2059, grant 62276078), the Key R\&D Program of Heilongjiang via grant 2022ZX01A32, the International Cooperation Project of PCL, PCL2022D01and the Fundamental Research Funds for the Central Universities (Grant No.HIT.OCEF.2023018).

\bibliography{latex/custom,latex/paper,latex/cr_new}

\appendix
\section{Appendix}
\subsection{Length Extrapolation on Generation Tasks}
\label{appdix:compa}
To help readers gain a deeper understanding of the challenges of length extrapolation, we leverage LongBench-E ~\citep{bai_longbench_2023} as our testbed and choose three trending LLMs with different context window sizes to evaluate their performance on various generation tasks and different evaluation length ranges. The results are shown in Table \ref{tab:compa}.

From the results, some intriguing conclusions can be drawn:
\begin{enumerate}
    \item When evaluating models on sequences beyond the original context window, a consistent performance degradation can be observed across models and tasks, which strongly supports the necessity of studying length extrapolation.
    \item Thanks to the shift-invariance and decay-with-distance property of RPE these LLMs use, they can maintain a reasonable performance when dealing with sequences beyond the context window, i.e., the performance will gradually decline rather than immediately crush after length exceeding the context window.
    \item Even evaluating on sequences within the context window, the increase in sequence length still leads to degraded performance. This may be as a result of the increasing difficulty with increasing length or due to the sparsity of long-range dependencies in concatenated training long sequences, meaning length extrapolation as a problem even exists within training context window and long-context transformers trained on long sequences do not necessarily possess strong length extrapolation capability.
\end{enumerate}

\begin{table*}
\centering
\setlength{\tabcolsep}{3pt}
\resizebox{\linewidth}{!}{
\begin{tabular}{lc|ccc}
\toprule
 Task& Evaluation Window & Llama2-7B-Chat (4K) &ChatGLM3-6B (8K) & Vicuna-v1.5-7b-16k \\
\midrule
\textit{\textbf{QA}} \\
\multirow{3}{*}{2WikiMQA}   & 0-4K & 34.56 & 21.86 & 31.19 \\
                            & 4-8K & 23.95 & 21.85 & 17.71 \\
                            & 8K+  & 23.12 & 13,72 & 12.33 \\
\hline
\multirow{3}{*}{HotpotQA}   & 0-4K & 37.59 & 25.92 & 37.35 \\ 
                            & 4-8K & 27.84 & 19.63 & 24.09 \\ 
                            & 8K+  & 23.17 & 15.96 & 21.91 \\ 
\hline
\multirow{3}{*}{MultiFieldQA-en}    & 0-4K & 41.42 & 44.04 & 47.1  \\ 
                                    & 4-8K & 34.29 & 29.31 & 33.83 \\ 
                                    & 8K+  & 21.21 & 28.45 & 28.29 \\
\midrule
\textit{\textbf{Summarization}} \\
\multirow{3}{*}{MultiNews}      & 0-4K & 26.67 & 25.71 & 27.96 \\
                                & 4-8K & 22.33 & 21.37 & 23.62 \\
                                & 8K+  & 22.46 & 20.4  & 21.22 \\
\hline
\multirow{3}{*}{GovReport}      & 0-4K & 30.66 & 30.7  & 33.95 \\ 
                                & 4-8K & 27.39 & 23.39 & 29.91 \\ 
                                & 8K+  & 25.6  & 22.2  & 24.89 \\
\midrule
\textit{\textbf{Code Completion}} \\
\multirow{3}{*}{LCC}        & 0-4K & 63.73 & 52.18 & 56.14 \\
                            & 4-8K & 61.59 & 43.63 & 57.69 \\
                            & 8K+  & 56.83 & 40.37 & 43.25 \\
\bottomrule
\end{tabular}}
\caption{Performance of Llama2-7B-Chat ~\citep{touvron_llama_2023-1}, ChatGLM3-6B ~\citep{glmChatGLMFamilyLarge2024} and Vicuna-v1.5-7b ~\citep{zhengJudgingLLMJudgeMTBench2023} on LongBench-E, where the context window of each model is indicated in parentheses.}
\label{tab:compa}
\end{table*}

\subsection{Results on Language Modeling}
\label{appdix:lm}
To offer an empirical comparison between popular PEs, we statistically collect results from published literatures and form Table \ref{tab:lm}.

We highlight several important conclusions from these results:
\begin{itemize}
    \item \textbf{RPEs demonstrate better in-distribution performance.} On sequences with length within context window, RPEs already demonstrate better performance, compared to APEs. We explain the results as RPE is consistent with the nature of natural language (relative position matters rather than absolute positions).
    \item \textbf{RPEs demonstrate better extrapolation capability.} In the length extrapolation setting that this survey concerns most, RPEs also outperform APEs due to intrinsic shift-invariance and binning strategy (for T5 bias) or exponentially decay with distance (for ALiBi and RoPE).
    \item \textbf{RPEs seek a balance between expressiveness (embedding-based RPE) and extrapolation (bias-based RPE) and perplexity is insufficient.} As in comparisons between RPEs, we can see that bias methods (T5 bias and ALiBi) lead to lower perplexity on sequences with length both within and beyond the context window, which indicates bias methods are better at language modeling by explicitly pandering recency bias. Note that it does not mean our claim that embedding-based methods like RoPE are more expressive is wrong, considering that models with ALiBi have worse performance than RoPE-based models on current trending benchmarks ~\citep{pal_giraffe_2023} like MMLU ~\citep{hendrycksMeasuringMassiveMultitask2020} and LMSys arena ~\citep{zhengJudgingLLMJudgeMTBench2023}. This further shows that perplexity is insufficient to reflect performance in these downstream tasks.
\end{itemize}

\begin{table*}
\centering
\setlength{\tabcolsep}{3pt}
\resizebox{\linewidth}{!}{
\begin{tabular}{l|cccc|cccc}
\toprule
 Dataset& \multicolumn{4}{c}{WikiText-103} & \multicolumn{2}{c}{OpenWebText2} & \multicolumn{2}{c}{ArXiv} \\
\midrule
 Context Window & \multicolumn{2}{c}{512} & \multicolumn{2}{c}{1024}& \multicolumn{4}{c}{512} \\
 Evaluation Window & 512 & 1012 & 1024 & 2024 & 512 & 1024 & 512 & 1024\\
\midrule 
 \textit{\textbf{APE}} \\
 Sinusoidal & 20.05 & 43.54 & 19.34 & 51.09 & 26 & 14168 & 5.8 & 1070 \\ 
 \textit{\textbf{RPE}} \\
 T5 Bias & 19.65 & 18.79 & 18.8 & 18.34 & 22.6 & 22.2 & 5.16 & 4.91 \\
 ALiBi & 19.73 & 18.73 & 18.66 & 18.05 & 22.8 & 23.3 & 5.25 & 5.41 \\
 RoPE & 20.07 & 21.37 & 19.33 & 31.17 & 23 & 61 & 5.25 & 16.02 \\ 
\bottomrule
\end{tabular}}
\caption{Empirical comparisons of different PEs on language modeling. The results on WikiText-103 are obtained from ~\citet{sun_length-extrapolatable_2023} and the results on OpenWebText2 and ArXiv are obtained from ~\citet{chiLengthExtrapolatableTransformers2024}. Note that the results may not be fairly comparable across dataset due to differences in model and training.}
\label{tab:lm}
\end{table*}

\subsection{Thoughts on Standardized Benchmark}
\label{appdix:benchmark}
Realizing the difficulty and complexity of constructing a standardized benchmark for length extrapolation, we present some preliminary thoughts on it as follows:
\begin{itemize}
    \item The benchmark should have \textbf{no position bias}. This means the model cannot consistently rely on tokens at specific locations to reach the correct answer. Thus, language modeling is not an ideal task due to its recency bias, which makes it possible for the model to generate the correct token based solely on nearby tokens.
    \item The benchmark should require \textbf{modeling the full range}. This indicates the model cannot depend on a small portion of the input but needs to attend and model the full range of context to give correct responses. Thus, the popular Needle In A Haystack test ~\citep{gkamradtGkamradtLLMTest_NeedleInAHaystack2024} is not an ideal benchmark, as it only requires the model to search and retrieve only a small portion of the input that is significantly different from other content, which is quite different from understanding and use of context ~\citep{liu_lost_2023}.
    \item This benchmark should offer \textbf{flexibility in sequence length with relatively stable difficulty}. This means the benchmark should consist of enough sequences at increasing lengths but not increasing difficulty. Thus, the benchmark can directly help with the fine-grained evaluation of the length extrapolation capability of Transformers without the need to crop a complete sequence, where the consistency of difficulty ensures the evaluation is only relevant to the increasing length.
\end{itemize}

As for a concrete example, calculating long sequences containing only addition and subtraction within ten (and keeping the intermediate results in a small range) might be a promising evaluation task, considering that the task itself is simple enough for common LLMs ~\citep{liCommon7BLanguage2024} and we can thus focus on the impact of increasing length.

\end{document}